%% file: main.tex
\documentclass[letterpaper]{article} 
\usepackage{aaai25}  
\usepackage{times}  
\usepackage{helvet}  
\usepackage{courier}  
\usepackage[hyphens]{url}  
\usepackage{graphicx} 
\usepackage{natbib}  
\usepackage{caption} 
\usepackage{algorithm}
\usepackage{algorithmic}

\usepackage{newfloat}
\usepackage{listings}
\DeclareCaptionStyle{ruled}{labelfont=normalfont,labelsep=colon,strut=off} 
\floatstyle{ruled}
\newfloat{listing}{tb}{lst}{}
\floatname{listing}{Listing}
\usepackage{array}
\newcolumntype{P}[1]{>{\centering\arraybackslash}p{#1}}

\usepackage{colortbl}
\usepackage{multirow}  
\usepackage{booktabs}
\usepackage{amsmath}
\usepackage{amsfonts}

\usepackage{array}
\newcolumntype{P}[1]{>{\centering\arraybackslash}p{#1}}

\usepackage{here}

\title{
    Limitations in Employing Natural Language Supervision for Sensor-Based Human Activity Recognition -- And Ways to Overcome Them.
}
\author{
    Harish Haresamudram\textsuperscript{\rm 1}, Apoorva Beedu\textsuperscript{\rm 1}, Mashfiqui Rabbi\textsuperscript{\rm 2}, \\Sankalita Saha\textsuperscript{\rm 2}, Irfan Essa\textsuperscript{\rm 1}, Thomas Pl\"otz\textsuperscript{\rm 1}
}
\affiliations{
    \textsuperscript{\rm 1}Georgia Institute of Technology \textsuperscript{\rm 2}Optum AI\\

    \{hharesamudram3, abeedu3, irfan.essa, thomas.ploetz\}@gatech.edu, \{mashfiqui.rabbi, sankalita.saha\}@optum.com
}

\begin{document}

\maketitle

\begin{abstract}
    Cross-modal contrastive pre-training between natural language and other modalities, e.g., vision and audio, has demonstrated astonishing performance and effectiveness across a diverse variety of tasks and domains. 
    In this paper, we investigate whether such natural language supervision can be used for wearable sensor based Human Activity Recognition (HAR), and discover that--surprisingly--it performs \textit{substantially worse} than standard end-to-end training and self-supervision.  
    We identify the primary causes for this as: \textit{sensor heterogeneity} and the \emph{lack of rich, diverse text descriptions of activities.}
    To mitigate their impact, we also develop  strategies and assess their effectiveness through an extensive experimental evaluation. 
    These strategies lead to significant increases in activity recognition, bringing performance closer to supervised and self-supervised training, while also enabling the recognition of unseen activities and cross modal retrieval of videos.
    Overall, our work paves the way for better sensor-language learning, ultimately leading to the development of foundational models for HAR using wearables.
\end{abstract}

%


\section{Introduction}
\input{latex_files/introduction.tex}

\section{Related Work}
\label{sec:related}
\input{latex_files/related_work.tex}

\section{Natural Language Supervision for HAR}
\label{sec:method}
\input{latex_files/method.tex}

\section{Experimental Settings}
\label{sec:settings}
\input{latex_files/settings.tex}

\section{Plug-and-Play NLS for HAR}
\label{sec:results}
\input{latex_files/results.tex}

\section{Going Beyond HAR through NLS}
\input{latex_files/analysis.tex}

\section{Summary and Conclusion}
\input{latex_files/conclusion.tex}
\clearpage

\appendix
\input{latex_files/appendix}

\clearpage

\bibliography{refs}
\clearpage

\end{document}

%% file: latex_files/introduction.tex

Learning joint embedding spaces by pairing modalities with natural language descriptions of their contents (e.g., image captions or sound descriptions for audio) has proven  successful across modalities \cite{radford2021learning, xu2021videoclip, wu2022wav2clip}.
Here, the task is to predict which description goes with which input, for large-scale datasets.
The expressiveness of natural language enables it to oversee a wider array of concepts, culminating in highly effective representations \cite{radford2021learning}.

Some advantages offered by such setups include: 
\emph{(i) zero-shot prediction of classes} 
through text descriptions, potentially aided by useful auxiliary and context information \cite{shen2022k}; 
and
\emph{(ii) cross-modal retrieval}, where natural language can be utilized to retrieve relevant images, audio, or sensor data, similar to performing search. 
The promise of these models is that they are \textit{plug-and-play}:  they can be utilized ``out-of-the-box'' in diverse applications, without requiring further training or adaptation. 

This paper focuses on natural language supervision (NLS) for Human Activity Recognition (HAR) using body-worn movement sensors, which involves automated prediction of what somebody is doing and when.
It has numerous applications, including health and fitness monitoring \cite{koskimaki2017myogym} and eating detection \cite{bin2022personalized}.
From a wearables standpoint, advantages of NLS such as predicting unseen activities and performing recognition through text queries are desirable, as they would allow new capabilities to be added on the fly. 

\begin{figure}[t]
    \centering
    \includegraphics[width=0.8\columnwidth]{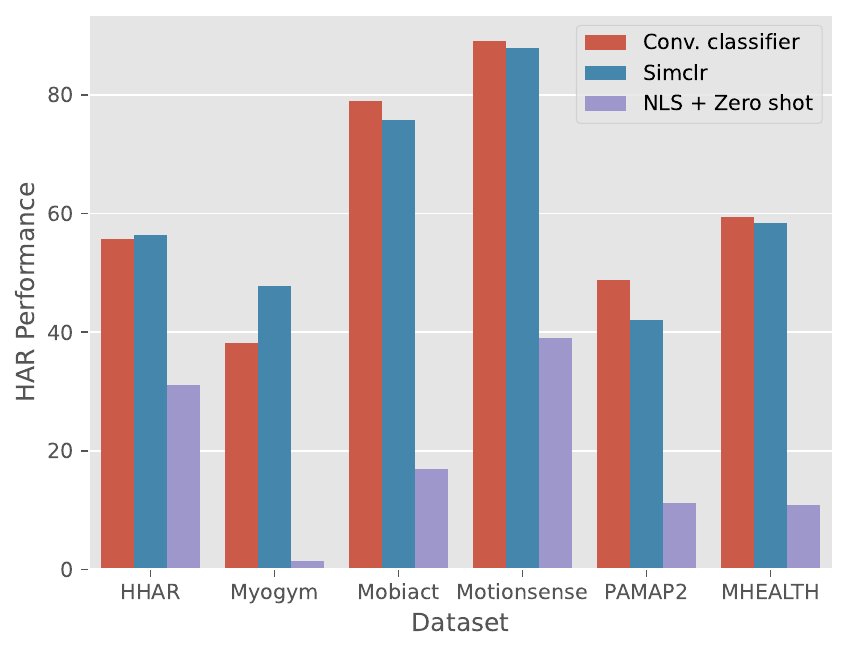}
    \caption{Difference in performance between supervised and self-supervised  training and natural language supervision.}
    \label{fig:intro_comparison}
\end{figure}

Despite the astonishing success of natural language supervision across modalities, domains, and applications, we discover and demonstrate in this paper that it is \textit{highly challenging} to apply NLS in a plug-and-play manner to wearable sensor (accelerometer) based HAR.
For example, Fig.\ \ref{fig:intro_comparison} shows that NLS based pre-training on a large-scale dataset (Capture-24) followed by zero shot prediction of activities is \textit{drastically worse} by around \textit{30-40\%}, than simple end-to-end and self-supervised training, across six target datasets. 

We discover two reasons that explain this reduction in performance, underpinned by challenges unique to wearable sensors and the HAR task:
\emph{(i) Sensor Heterogeneity:} Diversity in sensors results in significant differences in data distributions due to hardware constraints and settings such as gain, data and signal processing, differences in sampling rates -- even if the sensor locations and activities are the same \cite{stisen2015smart}.
This renders zero shot prediction very difficult, as pre-trained models cannot deal well with shifting distributions causing substantial performance degradation when there is no opportunity for adaptation to (vastly) different test conditions;
and
\emph{(ii) Lack of Rich Descriptions of Activities:} Learning such joint embedding spaces is data intensive, relying on diverse and unique text descriptions to learn wide ranging concepts. 
However, many HAR datasets contain only a handful of activity labels and (in some cases) demographics information \cite{kwon2020imutube, plotz2023if} -- a far cry from the 400M image-text pairs in the original CLIP paper \cite{radford2021learning}. 
Therefore, in scenarios with diverging data distributions, or when there is paucity of diverse descriptions of data, NLS can be an inferior option. 

We develop strategies to tackle these challenges, leading to improved HAR performance, and more broader applicability across scenarios. 
To deal with varying data distributions between pre-training and target datasets, we show how \textit{updating/adapting some layers of the pre-trained network on target data} with as little as 4 mins of data/activity leads to substantially improved recognition.
This demonstrates that adaptation with minimal amounts of target data can be sufficient for improved HAR, and potentially, across other applications with  distribution differences.
To improve diversity in text descriptions of activities, we not only explore the \textit{generation of additional prompts through LLMs}, but also study how \textit{external knowledge} can aid in improved recognition. 

The contributions of our work are as follows:
\begin{itemize}
    \item We adopt and adapt natural language supervision (NLS) for performing wearables-based HAR.
    
    \item We identify challenges rendering adaptation difficult and less successful, such as sensor heterogeneity and a lack of rich text descriptions accompanying sensor data.
    
    \item We develop strategies to tackle these challenges, enabling more successful application of NLS to sensor-based HAR, and potentially opening up such cross-modal training to other applications facing similar challenges.
    
\end{itemize}

%% file: latex_files/related_work.tex

Cross-modal contrastive training between natural language and other modalities has emerged as a highly effective training paradigm. 
Typically, the internet is crawled for collating large-scale datasets (of hundreds of millions of samples) with corresponding text descriptions.
Contrastive Language-Image Pre-training (CLIP  \cite{radford2021learning}), in particular, delivered superb performance across target scenarios by contrastively training to match images with corresponding text captions, and by using a 400M image-text pairs dataset. 
This training setup was subsequently adopted for video \cite{ma2022x, xu2021videoclip, zhao2023learning, bain2022clip} and audio \cite{elizalde2023clap, wu2022wav2clip, guzhov2022audioclip}, as well.
However, these methods require substantial training data containing rich and diverse text captions. 
When only keywords/class names are available, converting them to sentences for effective contrastive training has also been explored \cite{wu2023large}.
These methods are largely `plug-and-play', i.e., they can perform zero-shot recognition.

This setup has been extended to wearable sensors, through methods like IMU2CLIP \cite{moon2022imu2clip}, which uses the large-scale Ego4d dataset \cite{grauman2022ego4d}, containing  headmounted IMU and fine-grained text descriptions of activities.
As it was recorded at the head, it cannot be directly used for HAR, where common recording locations include the wrist or the waist.
ImageBind \cite{girdhar2023imagebind} also utilizes the Ego4d dataset to learn a joint embedding space between six modalities (incl. text and IMU) through contrastive pairwise training with vision as bridge.

For wireless sensors, TENT \cite{zhou2023tent} connects large language models (LLMs) to IoT sensors such as video, Radar, and LiDAR with text through a public dataset containing all modalities, and uses the pretrained CLIP text encoder for obtaining text embeddings.
More recently, Ts2Act \cite{xia2024ts2act} curated an image dataset for activity classes, and performed cross-modal contrastive pre-training with IMU data, for few shot recognition. 
Other works perform zero shot learning by learning to align IMU embeddings with synchronized video \cite{tong2021zero}.
The pre-trained video embeddings were utilized to learn a space to connect seen with unseen classes, leading to the ability to predict classes not seen for training.

Both Ts2Act and IMU2CLIP utilize different splits (i.e., train and test) from the same dataset, and therefore, do not perform HAR \textit{across datasets}.
Further, IMU2CLIP has access to rich text annotations whereas Ts2Act trains with images, which may be hard to obtain for rare activities.
Wearables datasets however, typically lack access to text data, beyond activity names. 
As such, our work is the first that focuses on directly pre-training with text sentences derived from activity labels.
This brings the capability to flexibly describe (in natural language) the movements present in activities, for HAR.
We also develop strategies to address challenges inherent to this setup, leading to the missing capability to function in truly plug-and-play fashion.

%% file: latex_files/method.tex
\begin{figure*}[!t]
    \centering
    \includegraphics[width=.9\textwidth]{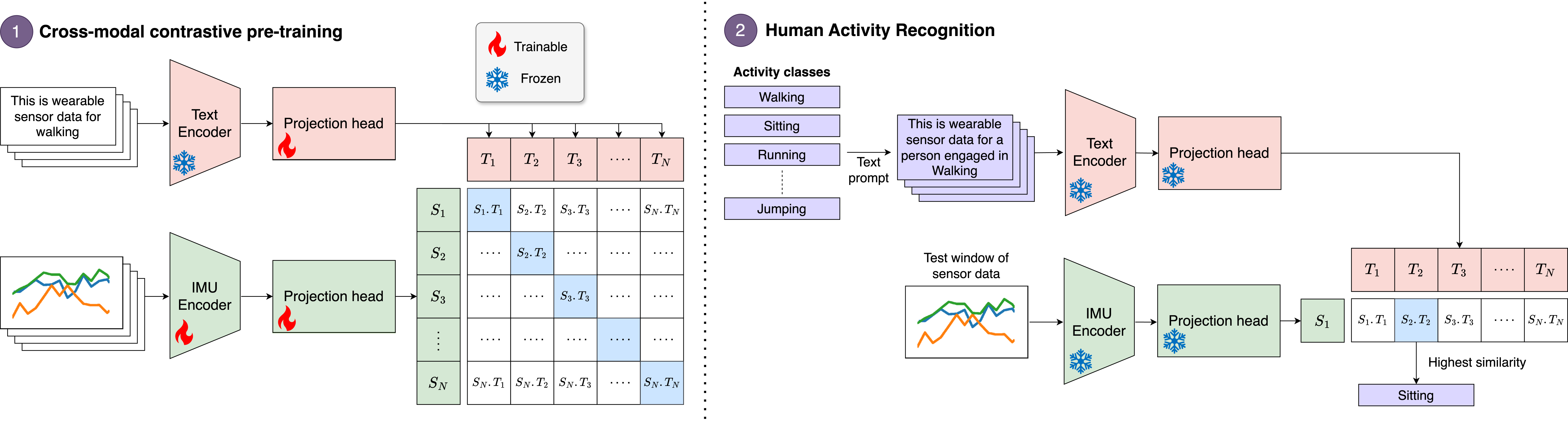}
    \caption{
    Natural language supervision for sensor-based HAR: the network is pre-trained by learning to accurately match windows of sensor data to the corresponding ground truth activities in form of textual descriptions. 
    HAR is then performed by computing cosine similarity scores between windows of test sensor data and all activity sentences.
    The sentence with highest similarity score determines the final activity output (lower right part in phase 2).
    This figure is inspired by  \cite{radford2021learning}. 
    }
    \label{fig:contrastive_sensor_language_framework}
\end{figure*}

Our NLS setup involves two stages: 
\textit{i)} cross-modal contrastive pre-training; and 
\textit{ii)} Human Activity Recognition. 

\subsection{Cross-Modal Contrastive Pre-Training}
Sensor data windows are first embedded through an encoder and a subsequent projection head, resulting in $N$ vectors $S=\{S_i\}_{i=1 \ldots N}$, where $N$ is the batch size. 
Correspondingly, the natural language descriptions of activities are also encoded through a text encoder and a separate projection head, to obtain text representations $T=\{T_i\}_{i=1 \ldots N}$.
Since only activity labels are available, simple text templates are employed to obtain sentences (Sec.\ \ref{sec:text_templates}).
Both $S_i$ and $T_i$ have dimension $D$.

In such a batch of $N$ sensor-text pairs, the task is to identify which of the $N \times N$ pairs are actual matches. 
We compute the cosine similarity ($C \in \mathbb{R}^{N \times N}$) for each $S_i$ and $T_i$, and train a joint embedding space to maximize the similarity of the $N$ actual pairs in the batch (i.e., the diagonal of the matrix in Fig.\  \ref{fig:contrastive_sensor_language_framework}), while minimizing the cosine similarity of the $N^2 - N$ incorrect pairings (i.e., the off diagonal elements in Fig.\  \ref{fig:contrastive_sensor_language_framework}).
In line with related work \cite{elizalde2023clap}, the similarity is computed as follows, where $\tau$ is the temperature parameter used to scale the logits:
\begin{equation}
    C = \tau * (S \cdot T^\top)
\end{equation}
\noindent
$\tau$ is initialized to $1/0.07$ and updated during training.

A symmetric cross entropy loss is optimized over these similarity scores, to update the network parameters \cite{elizalde2023clap, radford2021learning}:
\begin{equation}
    \mathcal{L} = 0.5 * (\ell_{\text{sensor}}(C) + \ell_{\text{text}}(C))
\end{equation}
\noindent
where, $\ell = \frac{1}{D} \sum^{D}_{i=0}\log(\text{diag}(\text{softmax}(C)))$, along the sensor and IMU axes respectively.

This setup contains three major components: 
\textit{(i) the IMU Encoder}, which provides embeddings of sensor data windows for contrastive pre-training. 
We utilize a convolutional encoder developed and utilized extensively in prior work \cite{saeed2019multi, haresamudram2022assessing};
\textit{(ii) the Text Encoder}, which is used to encode sentences of activities -- we employ the DistilBERT model \cite{sanh2019distilbert};
and
\textit{(iii) the Projection layers}, which are used to project embeddings from the modalities to a common representation space. 
In line with recent work, we use an MLP-based projection head. 
Detailed descriptions of the architecture are given in the Appendix (Sec.\ \ref{sec:architecture_details_appendix}).

\subsection{HAR Through Text-Based Classification}
\label{sec:text_templates}
HAR is performed by using sentences derived from activity labels (e.g., sitting, walking) and text templates (Fig.\  \ref{fig:contrastive_sensor_language_framework}). 
Unless specified differently we use the following, hand crafted template to obtain activity sentences: 
\texttt{
    \small
    This is wearable sensor data for a person engaged in \{activity\_name\}
}
where \texttt{\small activity\_name} is replaced with activities, e.g., walking, running, etc. 

For a target dataset containing $C$ classes we generate the sentences using the text template, and then compute the embeddings from the pre-trained text encoder and the learned projection head. 
Similarly, we compute the embeddings for the $N$ windows of target data using the learned encoder and projection heads.
As both embeddings are in a common space, we compute the cosine similarity between embeddings from each window of the target data (i.e., $N$ embeddings), and the embeddings for all target classes (i.e., $C$ embeddings).
The predicted label for each window is the class embedding with the highest cosine similarity, i.e., the class label is assigned based on the nearest  match to the window. 
This is shown in Part 2 of Fig. \ref{fig:contrastive_sensor_language_framework}.

%% file: latex_files/settings.tex
We study the effectiveness of NLS for HAR.
A sliding window approach is used to segment sensor data into overlapping windows, which are used for training and evaluation.
We now provide details of the HAR setup, the datasets, and the segmentation setup used for our experimental study.

\paragraph{Standard HAR setup}
\label{sec:standard_har_setup}
All target datasets are split into five folds by randomly sampling users, such that each user appears in the test set exactly once.
Similar to \cite{haresamudram2022assessing}, 20\% of the users are sampled randomly into the test set.
Of the remaining users, 20\% are sampled randomly into the validation set.
The rest of the users are used for training. 
There is \textit{no overlap of users} between the splits, yet the activities can be common.
For each fold, we normalize the train data to zero mean and unit variance, ditto for validation and test splits (using the statistics from the training data).
Average macro F1-scores across folds are reported for all experiments (unless specified differently).

\paragraph{Datasets}
For pre-training, we use the large-scale Capture-24 dataset \cite{willetts2018statistical}, which has 177 fine-grained labels after some data cleaning (correcting minor typos and removing semi-colons,  etc.), as they allow us to learn more concepts present in data. 
Following \cite{haresamudram2022assessing}, we evaluate on a diverse set of six target datasets, covering different recording locations (wrist, waist, ankle/leg), conditions, number of users, and types of activities.
Primarily, they contain locomotion-style activities (HHAR \cite{stisen2015smart}, Mobiact \cite{chatzaki2016human}, Motionsense \cite{malekzadeh2018protecting}, PAMAP2 \cite{reiss2012introducing}) and gym exercises (Myogym \cite{koskimaki2017myogym}, MHEALTH \cite{banos2014mhealthdroid}), in addition to daily living activities (PAMAP2). 
Details of the datasets are given in Tab.\ \ref{tab:datasets} in the Appendix.

\begin{table*}[!t]
    \centering
    \small
    \setlength{\tabcolsep}{1mm}
    \begin{tabular}{
                    P{.32\textwidth}  
                    P{.1\textwidth} 
                    P{.1\textwidth}  
                    P{.1\textwidth} 
                    P{.1\textwidth}  
                    P{.1\textwidth} 
                    P{.1\textwidth}}
    \toprule
    & 
    \multicolumn{2}{c}{Wrist} & 
    \multicolumn{2}{c}{Waist} & 
    \multicolumn{2}{c}{Leg} \\
    \cmidrule(lr){2-3} \cmidrule(lr){4-5} \cmidrule(lr){6-7}
    \multirow{-2}{*}{Method} & 
    HHAR & 
    Myogym & 
    Mobiact &             
    Motionsense & 
    MHEALTH & 
    PAMAP2 \\ 
	\midrule 
        \multicolumn{7}{c}{Baselines} \\ 
        \midrule
	Conv. classifier * & 55.63 $\pm$ 2.05 & 38.21 $\pm$ 0.62 & 78.99 $\pm$ 0.38 & 89.01 $\pm$ 0.89 & 48.71 $\pm$ 2.11 & 59.43 $\pm$ 1.56 \\ 
	DeepConvLSTM * & 52.37 $\pm$ 2.69 & 39.36 $\pm$ 1.56 & 82.36 $\pm$ 0.42 & 84.44 $\pm$ 0.44 & 44.43 $\pm$ 0.95 & 48.53 $\pm$ 0.98 \\ 
        \midrule
	Autoencoder + MLP classifier * &  53.64 $\pm$ 1.04 & 46.91 $\pm$ 1.07 & 72.19 $\pm$ 0.35 & 83.10 $\pm$ 0.60 & 40.33 $\pm$ 0.37 & 59.69 $\pm$ 0.72 \\ 
	SimCLR + MLP classifier * & 56.34 $\pm$ 1.28 & 47.82 $\pm$ 1.03 & 75.78 $\pm$ 0.37 & 87.93 $\pm$ 0.61 & 42.11 $\pm$ 0.28 & 58.38 $\pm$ 0.44 \\ 
	Enhanced CPC + MLP classifier * & 59.25 $\pm$ 1.31 & 40.87 $\pm$ 0.50 & 78.07 $\pm$ 0.27 & 89.35 $\pm$ 0.32 & 53.79 $\pm$ 0.83 & 58.19 $\pm$ 1.22 \\ 
        \midrule
        \multicolumn{7}{c}{Natural language supervision + zero shot prediction} \\ 
        \midrule
        NLS w/ pre-training on Capture-24 & 31.05 & 1.47 & 16.93 & 38.97 & 11.15 & 10.88 \\

        NLS w/ pre-train. on train split of target data & 29.05 & 33.30 & 59.09 & 73.36 & 41.72 & 48.36 \\
        
	\bottomrule
\end{tabular}
\caption{
    \textbf{HAR performance of natural language supervision (NLS) [mean F1]}: zero shot prediction is substantially worse than supervised and self-supervised baselines.  
    *: from \cite{haresamudram2023investigating} (five random classifier runs).
    }
\label{tab:out_of_domain_zero_shot}
\end{table*}

\paragraph{Sampling Rate and Segmentation}
For all experiments, we utilize raw accelerometer data, and downsample them  (if necessary) to 50Hz to match the lowest frequency across all datasets.
Following \cite{haresamudram2022assessing}, we set the sliding window size to 2 seconds, with 50\% overlap between segments.

%% file: latex_files/results.tex
\label{sec:challenges_with_nn_classification}

\begin{table}[!t]
    \centering
    \small
    \begin{tabular}{c c c P{0.4\columnwidth}}
        \toprule
        Dataset & \#classes & Vocab.\ size & Vocab.\  + template size \\
        \midrule
        Capture-24 & 176 & 282 & 292 \\
        HHAR & 6 & 9 & 19 \\
        Myogym & 31 & 55 & 65 \\
        Mobiact & 11 & 23 & 33 \\
        Motionsense & 6 & 8 & 18 \\
        PAMAP2 & 12 & 16 & 26 \\
        MHEALTH & 13 & 27 & 37\\
        \midrule
        ImageNet-21k* & $\sim$19.2k & 13.5k & -- \\
        YFCC 14M* & $\sim$14.2M & 2.41M & -- \\
        \bottomrule
    \end{tabular}
    \caption{Vocab.\ sizes of datasets. *: from \cite{shen2022k}}
    \label{tab:dataset_vocab_size}
\end{table}

The promise of NLS is that predictions can be made ``out-of-the-box''  through text queries without additional training, even on unseen classes.
We evaluate the effectiveness of NLS for HAR using the standard setup (Sec.\ \ref{sec:standard_har_setup}), and contrast its performance against supervised as well as self-supervised training with a large dataset (Capture-24) in Tab.\ \ref{tab:out_of_domain_zero_shot}. 

\subsection{Human Activity Recognition Experiments}
Here, we summarize the baselines of our evaluation (details in the Appendix), followed by a discussion of the results.

\paragraph{Baselines} 
\emph{(i) Supervised learning:} the Conv. classifier contains 1D convolutional layers whereas DeepConvLSTM \cite{ordonez2016deep} has 2D convolutional layers followed by an LSTM network.
These methods are trained end-to-end on the target datasets;
and 
\emph{(ii) Self-supervised learning:} the Autoencoder \cite{haresamudram2022assessing} is trained to reconstruct the input window after being passed through encoder and decoder layers. 
SimCLR \cite{tang2020exploring, chen2020simple} contrasts randomly augmented versions of the same input window whereas Enhanced CPC \cite{haresamudram2023investigating} performs contrastive training on future timesteps of sensor data.
They use 1D convolutional encoders along with an MLP classifier for HAR. 
Pre-training is performed on Capture-24 whereas the classifier layers are updated during classification on target data. 

\paragraph{Results}
Both end-to-end training and self-supervision substantially outperform NLS -- \textit{across datasets}.
In particular, NLS-based pre-training on Capture-24 and zero shot prediction on target datasets leads to very poor performance. 
This is because zero shot prediction involves \textit{no further training on target data.} 
Baselines perform substantially better than pre-training on Capture-24, as they have the chance to adapt to target conditions, by training at least some parts of the network with target data.
Similarly, pre-training on train splits of target datasets is also much better, as the model does not have to contend with (vastly) differing data distributions between training and testing. 
This leads to substantial increases of 30-50\%, yet under performing baselines. 
Overall, we find that zero shot prediction of activities for wearables based HAR is difficult and less effective than existing baselines.
Consequently, we conduct an analysis into the challenges affecting the performance of NLS.

\subsection{Challenges}
Differences in data distributions between pre-training and zero shot prediction have a substantial impact on performance.
We observe this clearly when pre-training on Capture-24, where there is no opportunity for the learned model to adapt to new target datasets, leading to a large performance drop.
Such differences in distributions are unique to wearables, where there is high diversity in sensors deployed, along with associated hardware constraints and settings (Fig.\  \ref{fig:data_distributions} in the Appendix shows details). 

The resulting diversity in recorded data, even for similar activities and movements, is referred to as \textit{`Sensor Heterogeneity'} \cite{koskimaki2017myogym}.
This is a \textit{challenge} and it prevents the use of large-scale pre-training, followed by zero shot prediction in diverse target conditions without adaptation -- in contrast to other domains, e.g., computer vision, where this is standard practice.

NLS benefits strongly from pre-training on large datasets containing \textit{rich and descriptive text} about diverse concepts, leading to generalization and effective zero shot prediction.
In contrast, most wearable datasets only contain activity names, thereby limiting what the model can learn.
Consequently, effectively learning a sensor-language joint embedding space is difficult, as observed in Tab.\ \ref{tab:out_of_domain_zero_shot}.
We summarize the number of classes and corresponding vocab.\ size used for pre-training in Tab.\ \ref{tab:dataset_vocab_size}, and observe that the vocab.\ size is 3-4 orders of magnitude smaller than for  vision datasets like ImageNet-21K \cite{deng2009imagenet} and YFCC-14M \cite{thomee2016yfcc100m} (as tabulated by \cite{shen2022k}).
As such, the \textit{lack of rich text descriptions of activities} is another challenge limiting performance of NLS. 

\section{Tackling NLS for HAR Challenges}
\label{sec:tackling}
Here, we detail strategies to tackle challenges in employing NLS for sensor-based HAR.
\subsection{Tackling Sensor Heterogeneity: Adapting Projection Layers on Target Data}
\label{sec:sub:adapting_proj}

Self-supervised methods shown in Tab.\ \ref{tab:out_of_domain_zero_shot} were also pre-trained on Capture-24, yet, training classifier layers with target data enables them to perform effective HAR.
Clearly,  learning / adapting at least some layers of the network (e.g., the classifier) on target data is necessary for usable HAR. 

We propose to update \textit{only the text and sensor projection heads} with the target labeled data (only the train split). 
This is an established practice in multi-modal setups, as it enables the alignment of modalities, especially when the encoders for different modalities are already pre-trained on different datasets \cite{moon2023anymal, verma2024mysterious}. 
We note that such an adaptation step, strictly speaking, violates the zero shot setup which does not allow further training.
Rather, it resembles few-shot learning, where small quantities of annotated target data can vastly improve HAR. 

\begin{figure}[t]
    \centering
    \includegraphics[width=0.9\columnwidth]{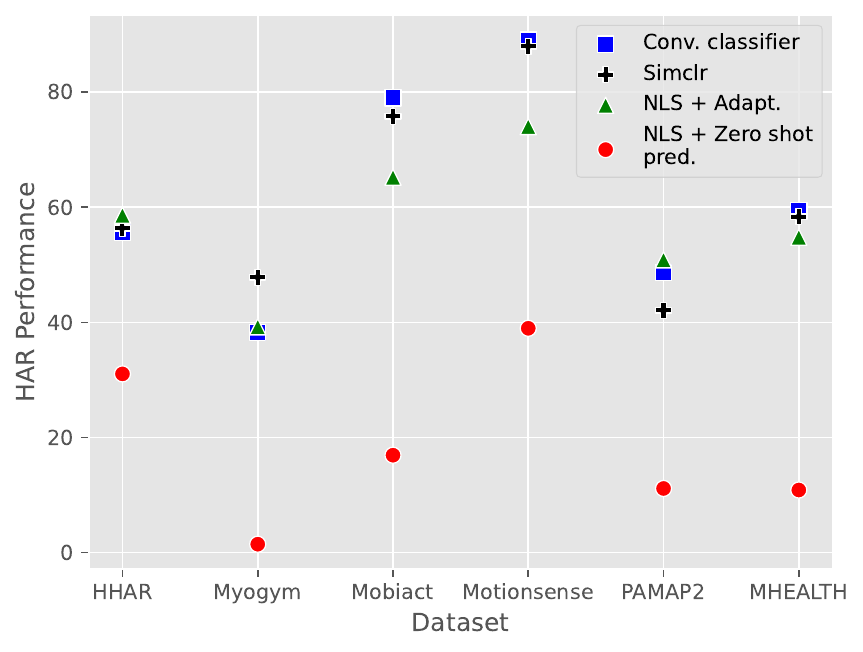}
    \caption{\textbf{Adapting projection layers} increases HAR performance of sensor-based NLS by 20-40\%. 
    }
    \label{fig:adaptation_full_data}
\end{figure}

For Fig.\ \ref{fig:adaptation_full_data}, we utilized the entire train split of target datasets for adaptation. 
Consistently, we observe \textit{substantial performance improvements resulting from adaptation}.
HHAR, Mobiact, MHEALTH, and PAMAP2 see increases of around 30-50\%, indicating the necessity of access to target annotations, bringing NLS closer to / beyond baselines.

\begin{figure}
    \centering
    \includegraphics[width=\columnwidth]{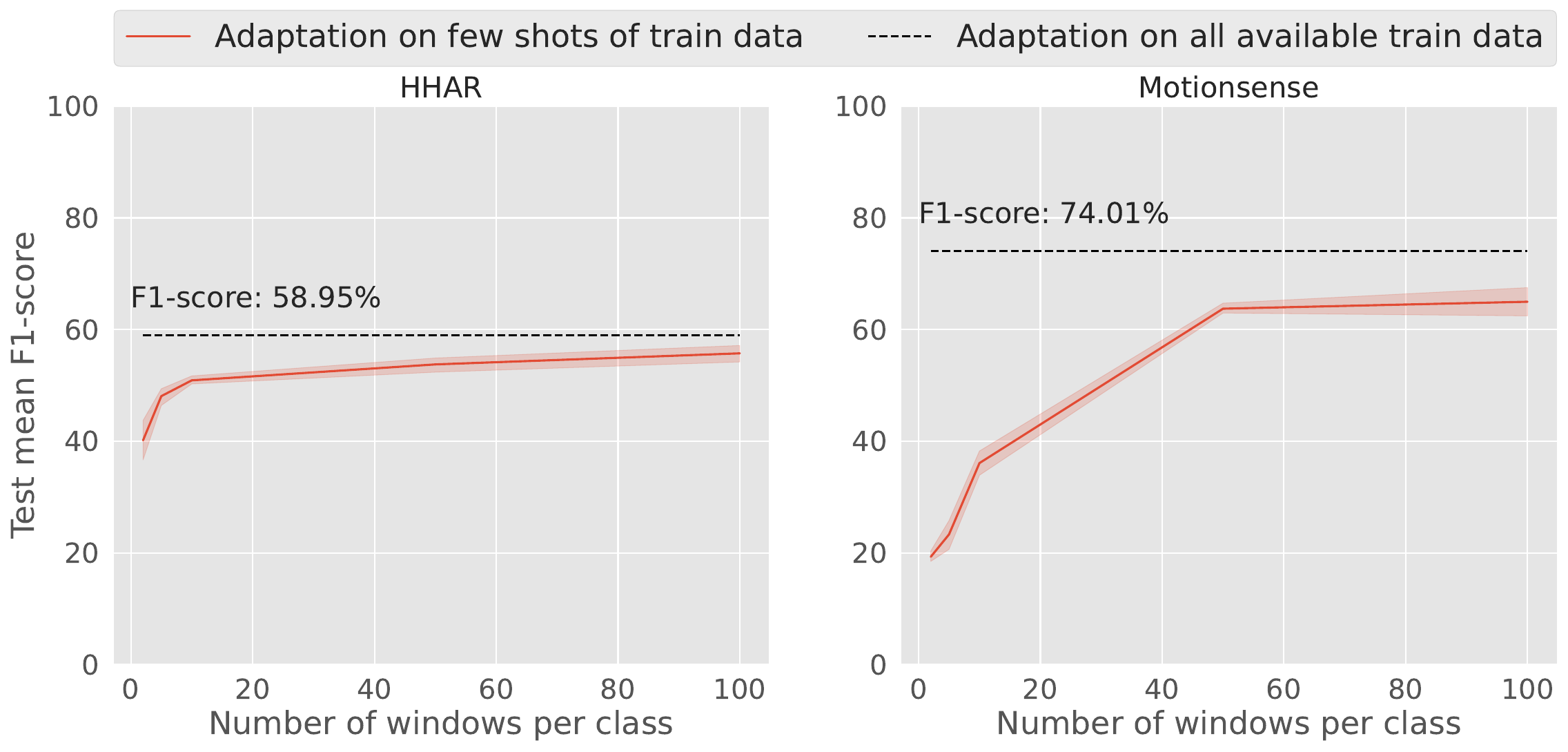}
    \caption{  
    \textbf{Adaptation on target data:} access to even small quantities of target data ($<$2 min) substantially improves performance.
    Full figure in the Appendix (Fig. \ref{fig:adaptation_few_shot}). 
    }
    \label{fig:adaptation_few_shot_small}
\end{figure}

For scenarios where it can be impractical to collect and annotate multiple hours of data, we also study if adaptation with small quantities of data can be useful.
To do so, we randomly sample $\{2,5,10,25,50,100\}$ windows/class for adaptation and report the performance on the test split, for five randomized runs in Fig.\ \ref{fig:adaptation_few_shot_small}.
Clearly, \textit{adapting projecting layers is highly advantageous}, with performance increases of 20-40\% (across datasets) with just 100 labeled windows, i.e., \textit{less than 4 minutes per activity}.

\begin{table*}[!t]
\centering
    \small
    \setlength{\tabcolsep}{1mm}
    \begin{tabular}{P{.01\textwidth}
                    P{.23\textwidth}
                    P{.19\textwidth}
                    P{.07\textwidth} 
                    P{.07\textwidth}  
                    P{.07\textwidth} 
                    P{.07\textwidth}  
                    P{.08\textwidth} 
                    P{.08\textwidth}}
    \toprule
    & & & 
    \multicolumn{2}{c}{Wrist} & 
    \multicolumn{2}{c}{Waist} & 
    \multicolumn{2}{c}{Leg} \\
    \cmidrule(lr){4-5} \cmidrule(lr){6-7} \cmidrule(lr){8-9}
    \# & Pretrain Setup & 
    HAR Setup & 
    HHAR & 
    Myogym & 
    Mobiact &             
    Mot.sense & 
    MHEALTH & 
    PAMAP2 \\ 
    \midrule
    1 & Base handcrafted template & Base handcrafted template & 29.05 & 33.30 & 59.09 & 73.36 & 41.72 & 48.36 \\
    \midrule
    2 & Randomly sample text template & Best template & 33.88 & 33.40 & 59.55 & 72.19 & 39.69  & 48.42 \\
    \midrule
    3 & Randomly sample ChatGPT diversified sentences & Best template  & 36.01 & \textbf{33.89} & 61.93 & 78.25 & \textbf{43.84} & 47.9 \\
    \midrule
    \midrule
    4 & Base handcrafted template + body parts info & Base handcrafted template + body parts info & 34.73 & 31.54 & 62.77 & 81.00 & 39.61  & 48.78 \\
    \midrule
    5 & Randomly sample text template + body parts info & Best template + body parts info & \textbf{36.43} & 32.80 & \textbf{65.68} & \textbf{82.93} & 40.36  & \textbf{51.90} \\
    \midrule
    6 & Randomly sample text template + movement descriptions & Best template + movement descriptions & 27.01 & 32.11 & 65.29 & 78.13 & 39.41  & 51.62 \\
    \bottomrule
\end{tabular}
\caption{
        \textbf{Increasing text diversity in activity descriptions:} Additional information about activities leads to better outcomes. 
    }
\label{tab:in_domain_more_diversity}
\end{table*}

\begin{table*}[!t]
    \centering
    \small
    \setlength{\tabcolsep}{1mm}
    \begin{tabular}{P{.42\textwidth}  
                    P{.30\textwidth}
                    P{.06\textwidth} 
                    P{.06\textwidth} 
                    P{.06\textwidth}}
    \toprule
    Method & Modifications &  HHAR & Mobiact & PAMAP2 \\ 
    \midrule 
    NLS w/ pre-training on train split of target data & -- & 29.05 & 59.09 & 48.36 \\
    \midrule

    NLS w/ pre-training on train split of target data  & Improved activity sentences + SLIP objective + CLIP text encoder & 50.2 & 63.69 & 51.95 \\

    \midrule
    NLS w/ pre-training on Capture-24 &  -- & 31.05 & 16.93 & 10.88 \\
    \midrule
    
    NLS w/ pre-training on Capture-24 + adapt.\ on target data & -- &  58.63 & 65.22 & 54.80 \\
    \midrule

    NLS w/ pre-training on Capture-24 + adapt.\ on target data & Improved activity sentences & 63.43 & 65.28 & 54.99 \\
\bottomrule
\end{tabular}
\caption{
    \textbf{Improving sensor-language systems:} Putting together findings from our exploration results in improved HAR. 
}
\label{tab:leveraging_practical_guidelines}
\end{table*}

\subsection{Tackling Lack of Rich Activity Descriptions: Increasing Text Diversity with LLMs} 
\label{sec:challenge_rich_descriptions}

We propose two measures to increase text diversity, and tabulate their impact in Tab.\ \ref{tab:in_domain_more_diversity}. 
For this setting, pre-training is performed on the train split of the target datasets, whereas the performance is reported on the test split:

\textit{(i) Additional text templates:} We hand-crafted eight text templates, and further employed ChatGPT to generate 25 additional (similar) templates, leading to a total of 33 templates for both pre-training and recognition. 
The underlying content of the sentences remains largely similar, while the activity information is presented in more diverse variations. 
As a result, any resulting performance improvements indicate the importance of text diversity.
We list the text templates in Tab.\ \ref{tab:list_of_templates} in the Appendix.

Similar to \cite{fan2023improving}, we also explore further diversification of the activity sentences (i.e., the sentences obtained after applying the template to  activities) via ChatGPT, by rewriting into 15 different variations.
Consequently, each of the 33 templates results in 15 variations per activity.

\noindent \textit{$\rightarrow$Insight: Diversity in the text descriptions is important}. 
From Tab. \ref{tab:in_domain_more_diversity}, we observe that increasing diversity through multiple text templates generally results in improved performance.
We observe only minor changes in recognition performance when using the base template vs.\ randomly sampling from the larger set.
However, utilizing ChatGPT to diversify activity sentences results in substantial boosts in performance across target datasets. 
For example, for Motionsense and HHAR, we see increases between 5-7\% over utilizing the base template (row 1 vs row 3 in Tab. \ref{tab:in_domain_more_diversity}). 

\textit{(ii) Leveraging external knowledge about activities: }
Additional context about classes, obtained through external knowledge sources, potentially enables improved generalization to \textit{new concepts} as they can be described using \textit{known concepts} \cite{shen2022k}. 
In our experiments, we utilize ChatGPT as the source of external knowledge, and query it to obtain the following information about activities: \textit{(i)} body parts; and \textit{(ii)} description of movements required, for performing the activities. 
For reference, Tab. \ref{tab:mobi_body_parts_info} and Tab. \ref{tab:mobi_movement_info} in the Appendix show external knowledge for Mobiact. \\
\textit{$\rightarrow$Insight: Adding external knowledge about activities is advantageous}.
Information about body parts and movements used for activities generally results in increased performance (rows 4-6 in Tab. \ref{tab:in_domain_more_diversity}). 
Even while using the base handcrafted template, HHAR, Mobiact, and Motionsense, observe improvements of around 6-10\% over the base setup. 
There are further increases when randomly sampling the set of templates during pre-training (i.e., from the 33 templates), and tuning for the best text template during activity recognition.

%% file: latex_files/analysis.tex
In Sec.\ \ref{sec:tackling}, we presented our experimental evaluation of NLS for standard wearables based HAR.
Here, we use NLS to go beyond, by exploring alternatives to components of the network, examining how our adapted NLS method recognizes unseen activities, and performing cross-modal search.

\subsection{Improving the NLS Setup}
We explore alternatives to components of the NLS setup, in order to further improve performance (full details in Sec.\ \ref{sec:investigating_components}):
\textit{(i) IMU Encoder:} Surprisingly, using simple convolutional encoders results in better performance than more complex ResNets (Fig.\ \ref{fig:diff_imu_encoders_all});
\textit{(ii) Text Encoder:} 
BERT \cite{devlin2018bert}, RoBERTa \cite{liu2019roberta}, and the CLIP Text encoder \cite{radford2021learning} perform better than DistilBERT, though they can have more parameters (Fig.\ \ref{fig:diff_text_encoders});
and 
\textit{(iii) Training Objective:} Advancements to the CLIP objective, e.g., allowing multiple matches between sensor windows and  sentences (UniCL \cite{yang2022unified}), and using an additional SimCLR loss (SLIP \cite{mu2022slip}) are advantageous (Fig.\ \ref{fig:diff_loss_functions_full}).
SLIP is the overall best option.

\textbf{Incorporating Better Alternatives:} We obtain substantial improvements in performance with the use of the aforementioned alternatives, as shown in Tab.\ \ref{tab:leveraging_practical_guidelines}. 
Using the CLIP text encoder and adding the SimCLR loss is a consistently superior option when pre-training on the train split of target datasets.
When studying adaptation on target data, using improved activity sentences (Sec.\ \ref{sec:challenge_rich_descriptions}) is also advantageous.

\subsection{Recognizing Unseen Activities}
\begin{table*}[!t]
	\centering
        \small
        \setlength{\tabcolsep}{1mm}
	\begin{tabular}{P{.38\textwidth}  
                    P{.09\textwidth} 
                    P{.09\textwidth}  
                    P{.09\textwidth} 
                    P{.09\textwidth}  
                    P{.09\textwidth} 
                    P{.09\textwidth}}
	\toprule
    & 
    \multicolumn{2}{c}{Wrist} & 
    \multicolumn{2}{c}{Waist} & 
    \multicolumn{2}{c}{Leg} \\
    \cmidrule(lr){2-3} \cmidrule(lr){4-5} \cmidrule(lr){6-7}
    Setup & 
    HHAR & 
    Myogym & 
    Mobiact &             
    Motionsense & 
    MHEALTH & 
    PAMAP2 \\ 
    \midrule
    Pre-training and HAR using the base text template & 55.01 & 36.96 & 56.71 & 40.68 & 39.54  & 55.33 \\
    \midrule
    Pre-training and HAR using base text template + body parts utilized & 33.16 & 19.96 & 64.01 & 51.01 & 30.33 & 53.73 \\
    \midrule
    Randomly sample text template during pre-training, and HAR with base template + body parts utilized & 44.82 & 28.85 & 54.46 & 63.86 & 41.13 & 54.02 \\
    \midrule
    Randomly sample text template during pre-training, and mean of all test sentence embeddings for HAR & 34.45 & 35.69 & 52.51 & 41.48 & 35.85 & 53.14 \\
    \bottomrule
\end{tabular}
\caption{
        \textbf{Recognizing unseen activities:} We obtain improved recognition on some datasets with addition of external knowledge.
 	}
\label{tab:zero_shot_recognition}
\end{table*}

\begin{figure*}[t]
    \centering
    \includegraphics[width=0.8\textwidth]{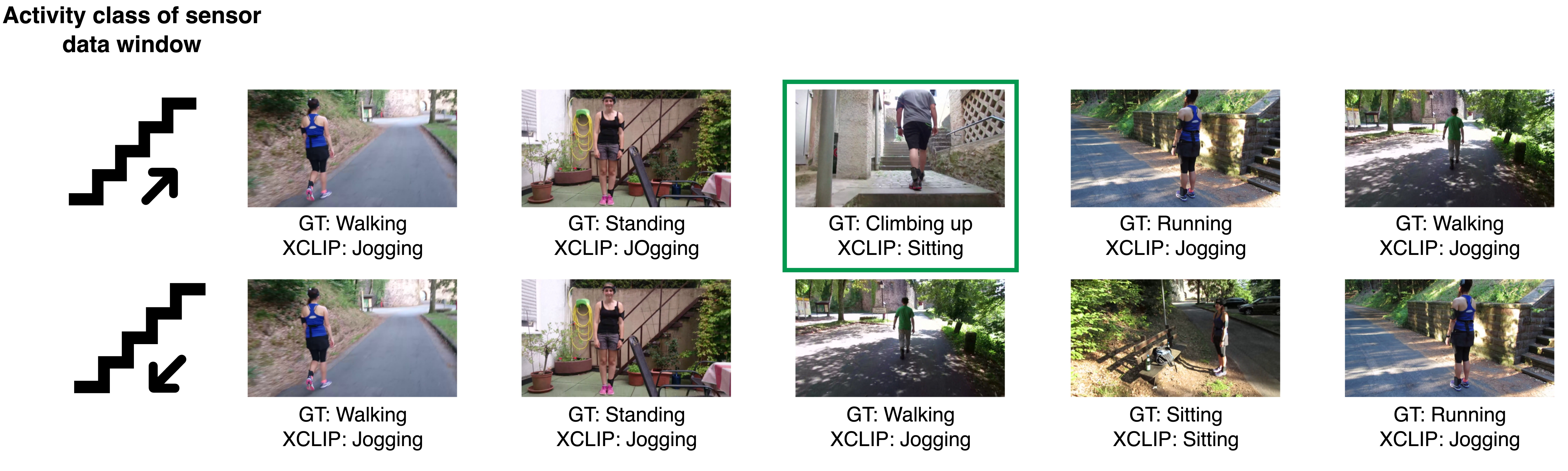}
    \caption{\textbf{Evaluating cross modal retrieval capabilities:} For four of the six activities from Motionsense, correct videos are retrieved among top-5 matches.
    `GT' is the ground truth label from RealWorld, whereas `XCLIP' comprises  predictions from the pre-trained X-CLIP model. 
    Full figure in the Appendix (see Fig.\ \ref{fig:cross_modal_retrieval})
    }
    \label{fig:cross_modal_retrieval_small}
\end{figure*}

A core advantage of NLS based classification is the ability to predict classes not seen during (pre-)training. 
To evaluate this capability, we utilize the protocol detailed in Sec.\ \ref{sec:zero_shot_setup}, where we partition datasets into 3-4 groups where \textit{the test set contains unseen activities}, which are non-overlapping across groups.
We tabulate results in Tab.\  \ref{tab:zero_shot_recognition}, and present the average of mean F1-scores obtained across groups.
We also modify the activity sentences based on the findings from Sec.\ \ref{sec:challenge_rich_descriptions}.

Solely using the base template performs well across the datasets, and is generally a good option.
Incorporating external knowledge from ChatGPT about the body parts involved is overall the best option, giving substantial increases throughout.
Similarly, adding descriptions of movements involved in activities is useful for some datasets.
This is in line with works from other domains such as computer vision  \cite{shen2022k}, where auxiliary information about the classes (e.g., specific birds having red feathers) typically results in performance increases under zero shot conditions. 
While natural language supervision for wearables-based HAR has challenges, the ability to perform zero shot prediction is a big plus for practical wearable systems.

\subsection{Cross Modal Retrieval of Videos}
We investigate whether we can use the text encoder from a pre-trained video-language model (X-CLIP \cite{ma2022x}) for contrastive training with sensor data, and subsequently \textit{retrieve} similar videos from a \textit{different dataset}, i.e., \textit{perform search.}
We evaluate on the \textit{RealWorld} dataset \cite{sztyler2016body}, which contains videos for locomotion-style activities, e.g., sitting, jogging.
For sensor-language pre-training, we use the X-CLIP text encoder and the \textit{Motionsense} dataset. 
We compute cosine similarity between sensor and video embeddings, and show the 5 closest video matches in Fig.\  \ref{fig:cross_modal_retrieval_small} for a random window per activity in Motionsense.
The setup is detailed in Sec.\ \ref{sec:cross_modal_retrieval_setup}.

For four classes--Walking up the stairs, Walking, Running, and Sitting--the correct video is retrieved among the top five matches.
On the other hand, the Standing and Sitting have many incorrect matches, usually with other static activities, where even supervised methods are routinely confused. 
Predictions by X-CLIP are often incorrect (despite being trained on large scale and diverse video-text datasets), and, consequently, the closest matches obtained after sensor-language training can also have reduced accuracy. 
In summary, this experiment shows that NLS aids in cross modal retrieval of videos, even though \textit{both sensor as well as the video encoders were not trained on same datasets. }

%% file: latex_files/conclusion.tex
We explored whether natural language supervision (NLS) can be employed for zero shot prediction of activities from sensor data in the typically promised plug-and-play manner.
We found that this is a \textit{very challenging} endeavor and identified its two primary causes: 
\textit{sensor heterogeneity}, and the \textit{lack of rich, diverse text descriptions of activities.}

Sensor heterogeneity causes HAR in diverging target conditions to be poor.
To tackle this, we proposed to use small amounts of labeled target data for adaptation.
To increase diversity in activity descriptions, we explored augmentation and incorporating external knowledge from pre-trained LLMs.
Both strategies resulted in substantial improvements. 
While sensor-language modeling does not outperform state-of-the-art supervised and self-supervised training for some datasets, its additional capabilities like recognizing unseen activities, and performing cross-modal search, are clearly advantageous for real world scenarios.
Our solutions result in improved sensor-language learning, paving the way for foundational models of human movements. 

%% file: latex_files/appendix.tex
\section{Appendix}

\subsection{Architecture Details}
\label{sec:architecture_details_appendix}

\begin{figure}[h]
    \centering
    \includegraphics[width=1\columnwidth]{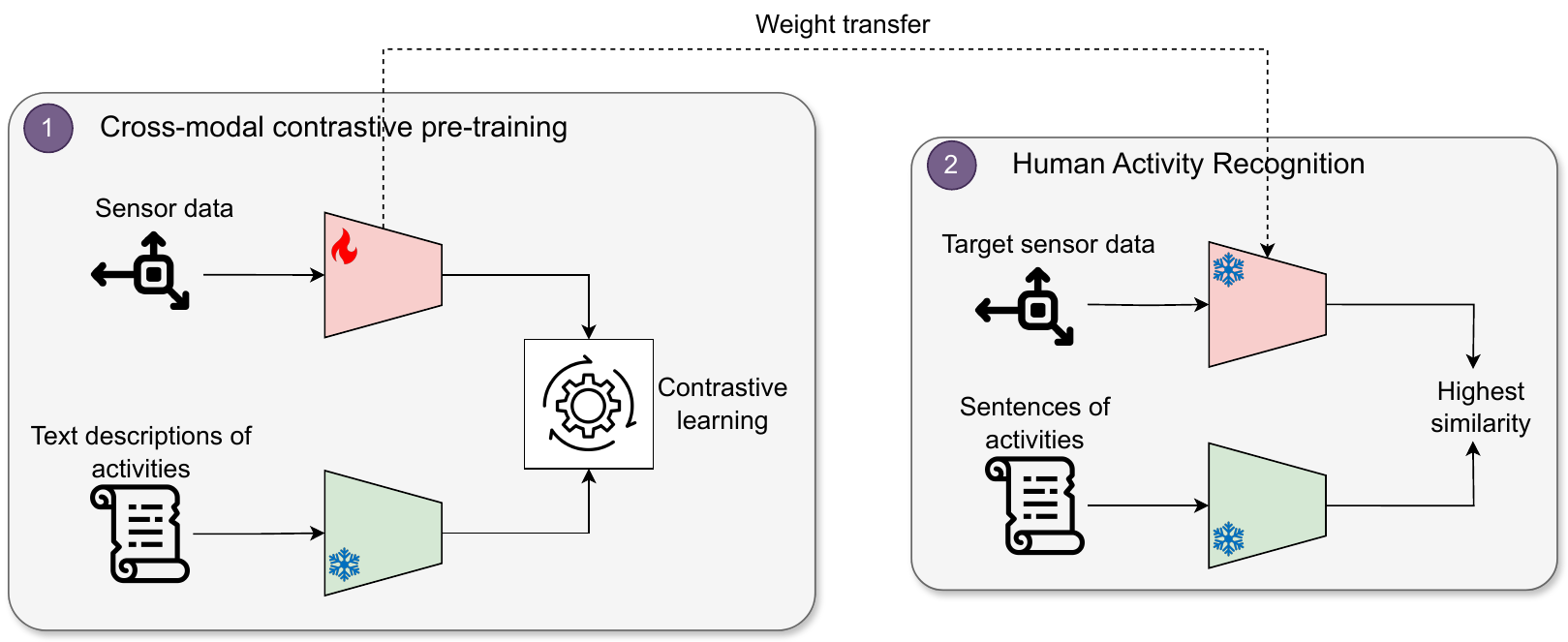}
    \caption{An overview of multi-modal contrastive learning. 
    }
    \label{fig:overview}
\end{figure}

The details of the architecture are as follows: 
\begin{enumerate}
    \item \textbf{IMU Encoder}: 
    It contains three blocks, each comprising a convolutional layer, ReLU activation, and dropout with p=0.2.
    The three convolutional layers have \{32, 64, 128\} filters and a filter size $3$, with reflect padding. 
    Max pooling is applied on the output of the last convolutional block, in order to obtain a single vector of size 128.
    Subsequently, a projection head is used to project to a higher dimensional vector, typically of dimension $512$ or $768$, to match the output of the text encoder. 

    \item \textbf{Text Encoder}: 
    Primarily, we utilize the DistilBERT \cite{sanh2019distilbert} for most of our experiments, as it is lightweight, while providing similar performance to the original BERT \cite{devlin2018bert}.
    We utilize the [CLS] token as the text encoder representation, which has a dimensionality of 768. 
    It has the same general architecture as BERT, but the number of layers is reduced by 2, in addition to the removal of the token-type embeddings and the pooler layers.
    Knowledge distillation is applied in order to learn the smaller sized student model, resulting in 97\% performance of the full BERT model, while being 40\% smaller in size and 60\% faster.
    We utilize the pre-trained models available on HuggingFace \cite{wolf2019huggingface} for all our experiments.
     
    \item \textbf{Projection layers}: 
    In the original formulation of the CLIP framework, a simple linear projection was applied to the encoder outputs, in order to compute the cosine similarity.
    More recently, more complex MLP-based projections have been investigated, leading to better performance \cite{elizalde2023clap, wu2023large}. 
    In line with this, we utilize two linear layers of 512 dimensions with ReLU in between.
\end{enumerate}

\subsection{Baselines}

\begin{enumerate}
    \item \textbf{Conv. classifier:} \cite{haresamudram2022assessing} we utilize an identical IMU encoder to Sec.\ \ref{sec:architecture_details_appendix}, and employ a three-layer MLP as the classifier. 
    The MLP contains (256, 128, and $num\_classes$) units respectively, with ReLU, dropout (p=0.2) and batch normalization in between. 
    The network is trained in a fully supervised manner.
    
    \item \textbf{DeepConvLSTM} \cite{ordonez2016deep}: sensor data are encoded with four 2D convolutional layers, each having 65 filters with a filter size of $5 \times 1$. 
    Subsequently, two LSTM layers of 128 units are utilized, followed by a linear layer.
    The entire network is trained end-to-end using annotations.
    
    \item \textbf{Autoencoder} \cite{haresamudram2022assessing}: 
    Autoencoders are trained by reconstructing input windows of sensor data after being passed through an encoder-decoder pair.
    The encoder has three 1D convolutional blocks, containing (32, 64, 128) filters with kernel size=3, and reflect padding. 
    After each convolutional layer, ReLU and dropout with p=0.2 is added. 
    The decoder is a mirror image of the encoder. 
    Mean Squared Error (MSE) loss between the input and reconstructed window of sensor data are utilized to update network parameters.
    During activity recognition, the learned encoder weights are frozen, and only the classifier layers are updated using cross entropy loss with target annotations. 
    The classifier is an MLP with three linear layers of (256, 128, and $num\_classes$) units, and ReLU, dropout (p=0.2) and batch normalization in between. 
    
    \item \textbf{SimCLR} \cite{chen2020simple, tang2020exploring}:
    SimCLR is trained by constrasting randomly augmented views of the same input windows against other pairs in the batch. 
    Identical to \cite{saeed2019multi}, the encoder has three 1D convolutional blocks, containing (32, 64, 96) filters with kernel sizes of (24, 16, 8) and ReLU and dropout (p=0.2) in between.
    The projection head is an MLP of three linear layers with (256, 128, 50) units, with ReLU in between. 
    The transformations are identical to \cite{tang2020exploring, um2017data} and all pairwise combinations are applied during pre-training. 
    Identical to the Autoencoder, only the classifier layers are updated during HAR. 
    Once again, the classifier is an MLP with three linear layers of (256, 128, and $num\_classes$) units, and ReLU, dropout (p=0.2) and batch normalization in between. 
    
    \item \textbf{Enhanced CPC} \cite{haresamudram2023investigating}: the underlying pretext task involves contrasting future timesteps of data against negatives derived from the batch.
    Windows of sensor data are encoded using a convolutional encoder containing four blocks, with (32,64,128,256) filters respectively, with kernel sizes of (4, 1, 1, 1) and stride of (2, 1, 1, 1).
    Each block comprises of a 1D convolutional layer, ReLU and dropout (p=0.2). 
    The encoded data are summarized using causal convolution blocks with 256 filters.
    The number of blocks and kernel sizes are tuned over (2, 4, 6) and (2, 3, 4, 5, 6, 7) respectively. 
    Identical to the Autoencoder, only the classifier layers are updated during HAR. 
    Once again, the classifier is an MLP with three linear layers of (256, 128, and $num\_classes$) units, and ReLU, dropout (p=0.2) and batch normalization in between.
    
\end{enumerate}

\subsection{Data Pre-Processing and Implementation Details}
\subsubsection{Processing Activity Labels into Sentences}
Unless specified differently, the template used to convert activity labels into sentences for pre-training and classification is: \texttt{\small This is wearable sensor data for a person engaged in \{activity\_name\}}, where \texttt{\small \{activity\_name\}} is replaced with activities, e.g., walking, running, etc. 
We also perform minor text cleanup of the fine-grained labels of Capture-24, including fixing typos, removing extra spaces, colons, etc.

\subsubsection{Implementation Details}
We use Pytorch \cite{paszke2019pytorch} for implementation.
For the LLMs, we utilize pre-trained models from HuggingFace \cite{wolf2019huggingface}. 
In order to pre-train on Capture-24, the hyperparameters are tuned over learning rate $\in \{1e-3, 1e-4, 5e-4\}$, weight decay $\in \{0, 1e-4\}$ and batch size $\in \{256, 512\}$.
When pre-training on train splits of the target datasets, the batch size is tuned over $\{128, 256\}$ instead, as they are substantially smaller than Capture-24.
We utilize the Adam optimizer and train for 50 epochs, with early stopping using patience of 5 epochs, based on the validation loss.
The parameters for adaptation are identical to the pre-training parameters.
Training is performed using a single Nvidia A40 GPU, on the Ubuntu 20.04 operation system.

\subsubsection{Metric}
Throughout, we utilize the mean F1-score (also called macro F1-score) as the metric. 
This is motivated by robustness of the mean F1-score to class imbalance, which is often present in wearable sensor datasets \cite{plotz2021applying}. 

\subsection{Experimental Setups}
\subsubsection{Setup for Recognizing Unseen Activities}
\label{sec:zero_shot_setup}
Here, activities not seen during training need to be directly recognized, and without any model training or adaptation.
Therefore, the dataset splitting involves creating held out test sets which contain only \textit{unseen classes}.
Consequently, the users can be common across the sets.

Similar to \cite{tong2021zero}, we create $k \in \{3,4,5\}$ test groups/sets by randomly sampling 2-5 activities (based on the number of total activities), such that each activity is contained in a test set only once.
For example, since Motionsense and HHAR contain 6 activities, we create three groups containing two activities each (randomly sampled, as mentioned earlier). 
As Mobiact, PAMAP2, and MHEALTH contain \{11, 12, 13\} activities respectively, we create four groups, containing four classes each whenever possible (the last group for Mobiact and MHEALTH contain three and five activities respectively). 
Myogym has 33 activities and we create six groups, comprising five activities for the first five groups, or three for the last group. 
The specific activities present in these groups are detailed in  Appendix \ref{sec:activity_splits_for_zsl}.

We perform minimal hyperparameter tuning for this experiment, in line with current zero shot learning literature \cite{tong2021zero}.
Specifically, we aim to avoid violating the zero shot assumption, as our test set contains unseen classes.
As a result, we used training parameters that are commonly used in the literature for   pre-training (the classification does not have any hyperparameters): a learning rate of $1e-4$, weight decay of $0$, batch size of $256$, and the number of epochs is $50$.
In addition, we also emphasize that performance on the zero-shot setup cannot be directly compared to standard HAR, as not only are some classes unseen, but the similarity computation is performed against \textit{only} the unseen classes. 

\subsubsection{Setup for Cross Modal Retrieval of Videos Using Sensor Data}
\label{sec:cross_modal_retrieval_setup}
We use the pre-trained X-CLIP text encoder and projection layers, and freeze their weights. 
Only the IMU encoder and projection layers are updated during sensor-language pre-training with the \textit{Motionsense dataset}.
From each video in Realworld, we randomly sample one second segment from the first 15 seconds of the video, and compute the video embeddings using the pre-trained X-CLIP video encoder (which uses the first eight frames from the sampled segment).
Subsequently, we compute the cosine similarity of sensor embeddings to the video embeddings and display the 5 closest video matches in Fig. \ref{fig:cross_modal_retrieval} for a randomly sampled window from each activity in Motionsense.
We employ the base text template for both pre-training and HAR.

\subsection{Visualizing the data distributions of wearable sensor datasets}
\begin{figure}[h]
    \centering
    \includegraphics[width=1\columnwidth]{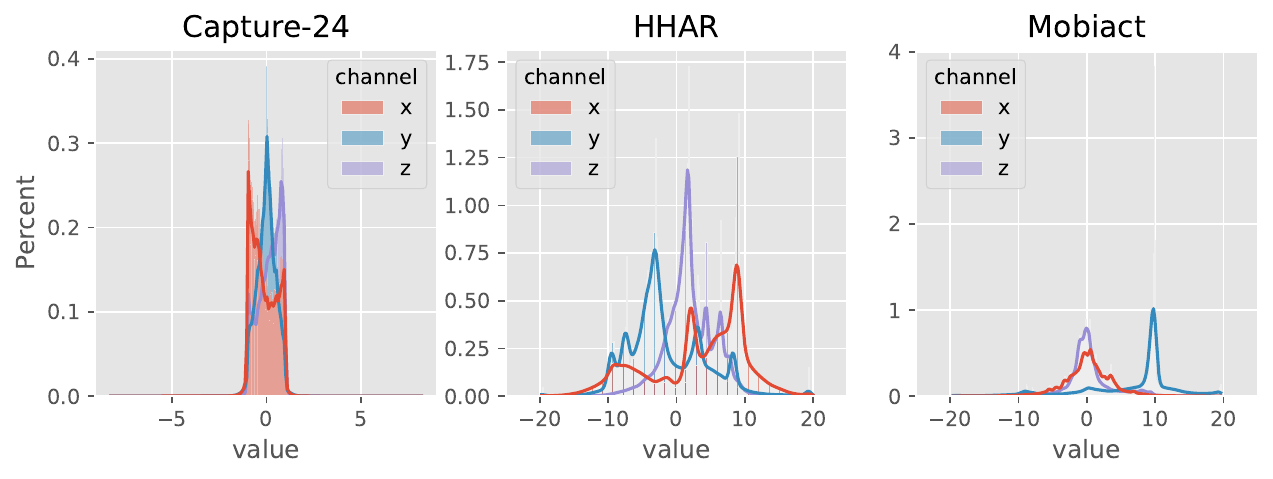}
    \caption{
    Visualizing the data distributions of various datasets: we observe that there are substantial differences across target datasets. 
    Further, we hypothesize that this results in worsened performance for HAR, if there is no access to target data for adaptation.
    }
    \label{fig:data_distributions}
\end{figure}

\null\clearpage

\subsection{Activity Splits for Zero Shot Recognition}
\label{sec:activity_splits_for_zsl}
\begin{table}[h]
    \centering
    \begin{tabular}{cc}
        \toprule
        Group & Activities \\
        \midrule
        1 & Going down the stairs, Walking \\
        2 & Biking, Going up the stairs \\
        3 & Sitting, Standing \\
        \bottomrule
    \end{tabular}
    \caption{Data preparation for HHAR for zero shot recognition of activities}
    \label{tab:zero_shot_hhar}
\end{table}

\begin{table}[h]
    \centering
    \begin{tabular}{c P{7.0cm}}
        \toprule
        Group & Activities \\
        \midrule
        1 & Standing, Nordic walking, Ascending stairs \\ 
        2 & Cycling, Descending stairs, Rope jumping \\ 
        3 & Sitting, Walking, Ironing \\ 
        4 & Sleeping, Running, Vacuum cleaning \\
        \bottomrule
    \end{tabular}
    \caption{Data preparation for PAMAP2 for zero shot recognition of activities}
    \label{tab:zero_shot_pamap}
\end{table}

\begin{table}[h]
    \centering
    \begin{tabular}{c P{7.0cm}}
        \toprule
        Group & Activities \\
        \midrule
        1 & Jogging, Transition from standing to sitting, Transition from sitting to standing \\
        2 & Walking down the stairs, Sitting on a chair, Stepping out of a car \\
        3 & Standing, Walking, Continuous Jumping \\
        4 & Walking up the stairs, Stepping into a car \\
        \bottomrule
    \end{tabular}
    \caption{Data preparation for Mobiact for zero shot recognition of activities}
    \label{tab:zero_shot_hhar}
\end{table}

\begin{table}[h]
    \centering
    \begin{tabular}{c  P{7.0cm}}
        \toprule
        Group & Activities \\
        \midrule
        1 & Walking up the stairs, Walking \\
        2 & Walking down the stairs, Sitting \\
        3 & Jogging, Standing \\
        \bottomrule
    \end{tabular}
    \caption{Data preparation for Motionsense for zero shot recognition of activities}
    \label{tab:zero_shot_hhar}
\end{table}

\begin{table}[h]
    \centering
    \begin{tabular}{c  P{7.0cm}}
        \toprule
        Group & Activities \\
        \midrule
        1 & Sleeping, Frontal elevation of arms, Running \\ 
        2 & Sitting, Waist bends forward, Knees bending (crouching) \\ 
        3 & Climbing stairs, Jogging, Jump front and back \\ 
        4 & Transitions between activities, Standing still, Walking, Cycling \\
        \bottomrule
    \end{tabular}
    \caption{Data preparation for MHEALTH for zero shot recognition of activities}
    \label{tab:zero_shot_hhar}
\end{table}

\begin{table}[h]
    \centering
    \begin{tabular}{c P{7.0cm}}
        \toprule
        Group & Activities \\
        \midrule
        1 & One Arm Dumbbell Row, Wide Grip Pulldown Behind The Neck, Reverse Grip Bent-Over Row, Bench Press, Dumbbell Alternate Bicep Curl \\
        2 & Seated Cable Rows, Leverage Chest Press, Close Grip Barbell Bench Press, Incline Hammer Curl, Front Dumbbell Raise \\
        3 & Wide Grip Front Pulldown, Pushups, Bar Skullcrusher, Tricep Dumbbell Kickback, Spider Curl \\
        4 & Dumbbell Flyes, Concentration Curl, Cable Curl, Hammer Curl, Seated Dumbbell Shoulder Press \\
        5 & Incline Dumbbell Flyes, Bench Dip, Upright Barbell Row, Side Lateral Raise, Lying Rear Delt Raise \\
        6 & Transition between activities, Bent Over Barbell Row, Incline Dumbbell Press, Triceps Pushdown, Overhead Triceps Extension, Car Drivers \\
        \bottomrule
    \end{tabular}
    \caption{Data preparation for Myogym for zero shot recognition of activities}
    \label{tab:zero_shot_hhar}
\end{table}

\clearpage

\subsection{External Knowledge about the Body Parts and Movements Involved in Performing Activities}

\begin{table}[h]
    \centering
    \small
    \begin{tabular}{P{2cm} P{6cm}}
        \toprule
        Activity & Body parts involved \\
        \midrule
        Standing & Primarily utilizes the muscles in the legs and core to maintain an upright posture. \\ 
        Walking & Involves the legs, hips, and feet, with arm and core engagement for balance and momentum. \\ 
        Jogging & Uses leg muscles, hips, feet, core, and arms for movement and maintaining pace. \\ 
        Continuous Jumping & Engages leg muscles, particularly calves and thighs, along with core muscles for stability. \\ 
        Walking up the stairs & Employs the quadriceps, hamstrings, glutes, calf muscles, and core for lifting the body upwards. \\ 
        Walking down the stairs & Utilizes the quadriceps, hamstrings, and core muscles to control the descent. \\ 
        Transition from standing to sitting & Involves hip flexors, quadriceps, and glute muscles to lower the body into a seated position. \\ 
        Sitting on a chair & Relies on the muscles of the buttocks, thighs, and back to maintain a seated posture. \\ 
        Transition from sitting to standing & Engages the quadriceps, glutes, hamstrings, and core to rise to an upright stance. \\ 
        Stepping into a car & Requires the use of leg muscles, especially the hip flexors, and core for balance. \\ 
        Stepping out of a car & Involves the legs, particularly the quadriceps and hamstrings, and core muscles for stability while exiting. \\
        \bottomrule
    \end{tabular}
    \caption{
    Information about the body parts involved while performing activities from the Mobiact dataset. Obtained from ChatGPT (model=GPT4.0).
    For brevity, we present external knowledge generated by ChatGPT only for activities from the Mobiact dataset. 
    }
    \label{tab:mobi_body_parts_info}
\end{table}
\null

\begin{table}[h]
    \centering
    \small

    \begin{tabular}{P{2cm} P{6cm}}
        \toprule
        Activity & Movements involved \\
        \midrule
        Standing &  Maintaining an upright position without significant movement. \\ 
        Walking &  Rhythmic stepping with alternating leg movements, causing a periodic change in body position and displacement over time. \\ 
        Jogging &  A form of trotting or running at a slow or leisurely pace, involving a bounce in step and increased heart rate. \\ 
        Continuous Jumping &  Repeatedly propelling oneself off the ground using both feet and landing with both feet, typically in the same spot. \\ 
        Walking up the stairs &  Ascending a set of steps by stepping up with alternating feet, often involving a shift in weight and balance. \\ 
        Walking down the stairs &  Descending a set of steps by stepping down with alternating feet, requiring controlled movement and balance. \\ 
        Transition from standing to sitting &  Lowering the body from an upright position to a seated position, typically involving bending at the knees and hips. \\ 
        Sitting on a chair &  Being stationary with the body supported by the buttocks and thighs, and the torso usually upright, while resting on a chair. \\ 
        Transition from sitting to standing &  Raising the body from a seated position to an upright stance, involving straightening the knees and hips. \\ 
        Stepping into a car &  Lifting one leg and then the other to move horizontally into the vehicles cabin, typically bending the torso and often using the hands for support. \\ 
        Stepping out of a car &  Moving one leg and then the other from the vehicles cabin to the ground, typically involving a pivot and a shift in balance to exit. \\
        \bottomrule
    \end{tabular}
    \caption{
    Information about the movements involved while performing activities from the Mobiact dataset. Obtained from ChatGPT (model=GPT4.0).
    For brevity, we present external knowledge generated by ChatGPT only for activities from the Mobiact dataset. 
    }
    \label{tab:mobi_movement_info}
\end{table}

\clearpage

\subsection{Improving the Natural Language Supervision Setup for HAR}
\label{sec:investigating_components}
\medskip
\begin{minipage}{\textwidth}

    \centering
    \includegraphics[width=\textwidth]{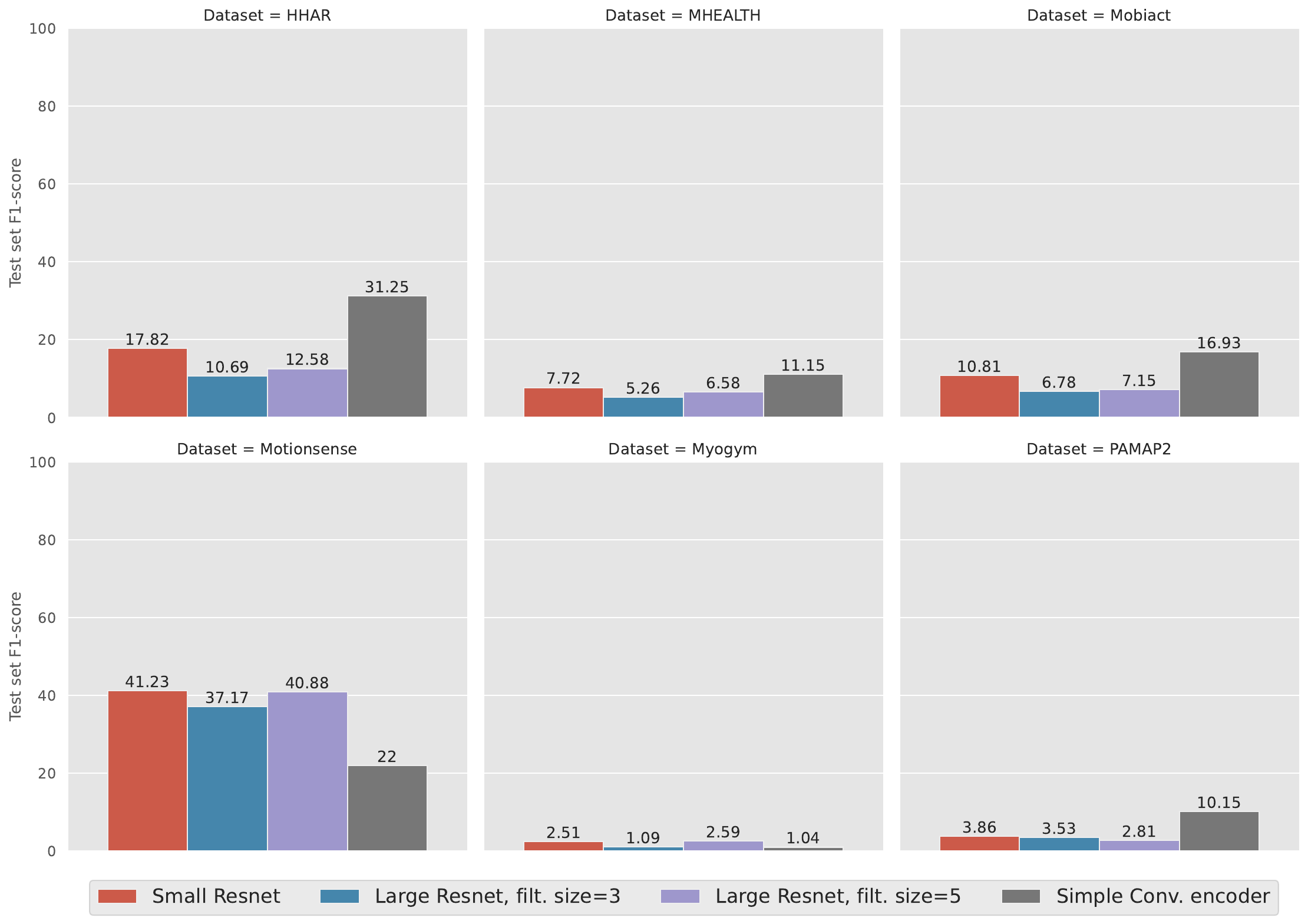}
    \captionof{figure}{Impact of the IMU encoder architecture used during pre-training.}
    \label{fig:diff_imu_encoders_all}
\medskip
    \centering
    \includegraphics[width=\textwidth]{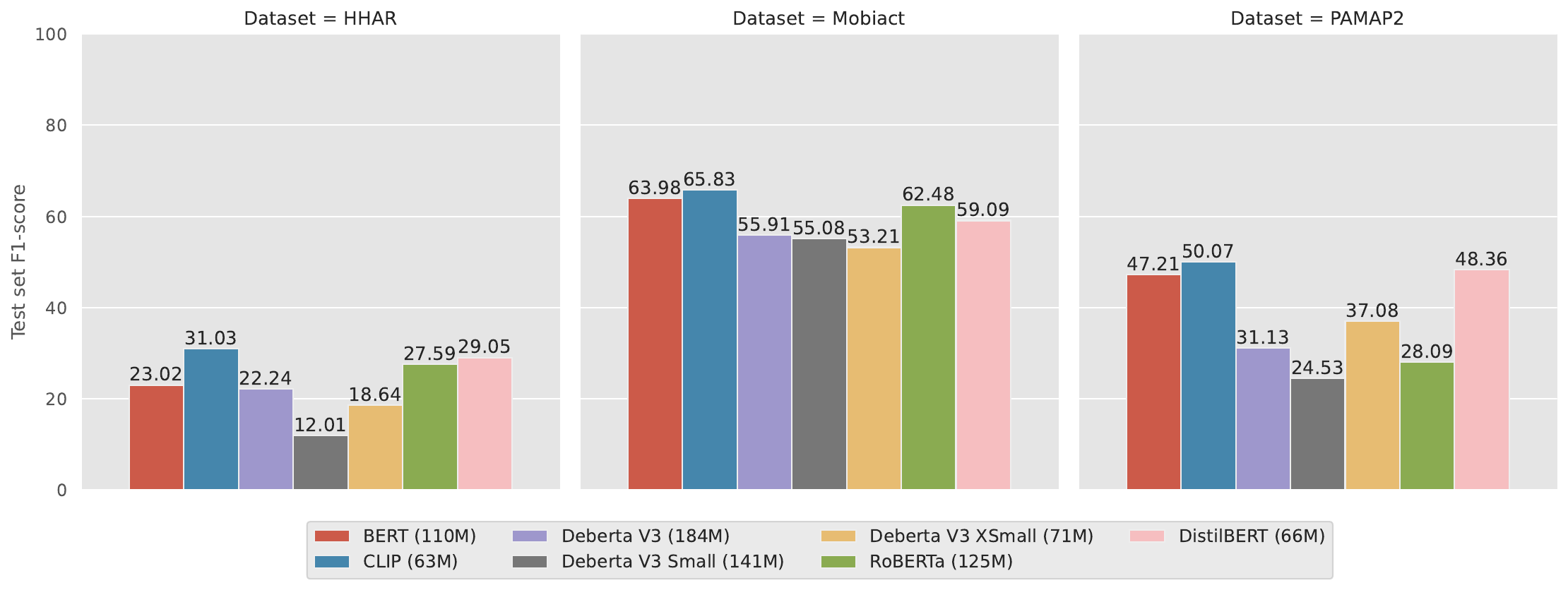}
    \captionof{figure}{Studying the impact of the language model used for encoding text: we observe that CLIP and DistilBERT (which are the smallest encoders), are the most effective options.
    We show the approximate number of parameters in these models in brackets, e.g., BERT has approximately 110M parameters.
    }
    \label{fig:diff_text_encoders}
\end{minipage}

\clearpage

\begin{figure*}[t]
    \centering
    \includegraphics[width=0.8\textwidth]{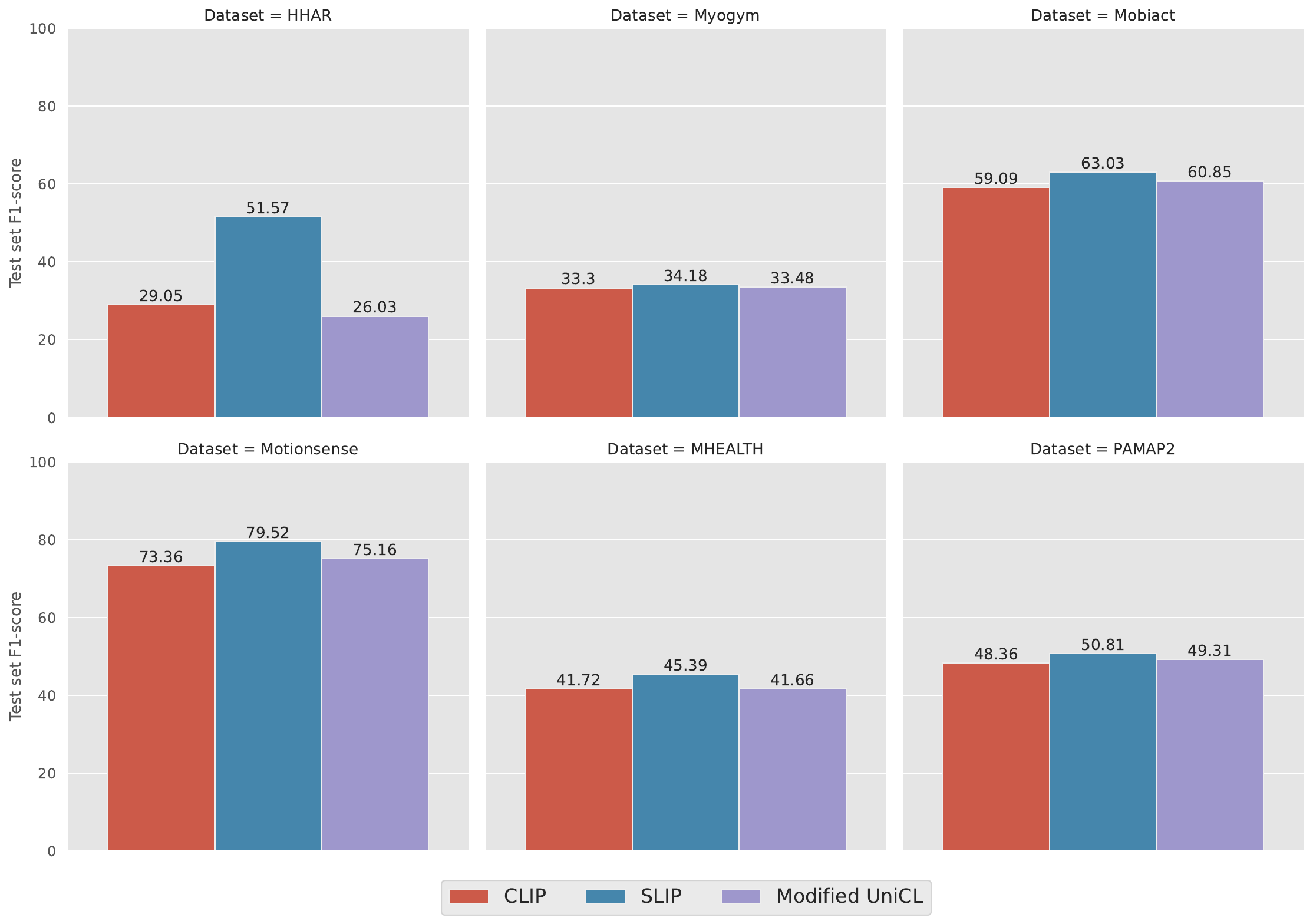}
    \caption{Impact of the training objective used during pre-training.}
    \label{fig:diff_loss_functions_full}
\end{figure*}
\vfill

\clearpage

\subsection{Summary of Datasets}
\medskip
\begin{minipage}{\textwidth}
\centering
\small
\begin{tabular}{P{3.0cm}|P{1.5cm}|c|c|P{10cm}}
    \toprule
    Dataset & Location & \# Users & \# Act. & Activities \\ 
    \midrule
    Capture-24 \cite{chan2021capture, gershuny2020testing, willetts2018statistical} & Wrist & 151 & 177 & Fine-grained free living activities. E.g., carrying heavy loads, home activity child elderly pet care, home activity child elderly pet care child care child care sitting kneeling, home activity household chores grocery shopping shopping, home activity household chores house cleaning,
    leisure sports miscellaneous hiking or walking at a normal pace through fields and hillsides, manual work, ... \footnote{Obtained after minor text cleaning.}
    \\ 
    \hline
    HHAR \cite{stisen2015smart} & Wrist & 9 & 6 & Biking, sitting, going up and down the stairs, standing, and walking \\ 
    \hline
    Myogym \cite{koskimaki2017myogym} & Wrist & 10 & 31 & Seated cable rows, one-arm dumbbell row, wide-grip pulldown behind the neck, bent over barbell row, reverse grip bent-over row, wide-grip front pulldown, bench press, incline dumbbell flyes, incline dumbbell press and flyes, pushups, leverage chest press , close-grip barbell bench press, bar skullcrusher, triceps pushdown, bench dip, overhead triceps extension, tricep dumbbell kickback, spider curl, dumbbell alternate bicep curl, incline hammer curl, concentration curl, cable curl, hammer curl, upright barbell row, side lateral raise, front dumbbell raise, seated dumbbell shoulder press, car drivers, lying rear delt raise, null \\ 
    \midrule
    Mobiact \cite{chatzaki2016human} & Waist/ Trousers & 61 & 11 & Standing, walking, jogging, jumping, stairs up, stairs down, stand to sit, sitting on a chair, sit to stand, car step-in, and car step-out\\ 
    \hline
    Motionsense \cite{malekzadeh2018protecting} & Waist/ Trousers & 24 & 6 & Walking, jogging, going up and down the stairs, sitting and standing \\ 
    \midrule
    MHEALTH \cite{banos2014mhealthdroid} & Leg/ Ankle & 10 & 13 & Standing, sitting, lying down, walking, climbing up the stairs, waist bend forward, frontal elevation of arms, knees bending, cycling, jogging, running, jump front and back, null class \\ 
    \hline
    PAMAP2 \cite{reiss2012introducing} & Leg/ Ankle & 9 & 12 & Lying, sitting, standing, walking, running, cycling, nordic walking, ascending and descending stairs, vacuum cleaning, ironing, rope jumping \\ 
    \bottomrule
    \end{tabular}
    \captionof{table}{
    A summary of the datasets used in our evaluation: we utilize the fine-grained labels from Capture-24 for generating the text descriptions used for contrastive training. 
    We study six target datasets, using their ground truth classes for HAR. 
    This table has been adopted with permission from \cite{haresamudram2022assessing}.
}
\label{tab:datasets}
\end{minipage}

\null\clearpage

\subsection{Text Templates Used for Pre-training and HAR}
\medskip
\begin{minipage}{\textwidth}
    \centering
    \small
    \begin{tabular}{P{0.15\textwidth}|P{0.75\textwidth}}
        \toprule
        Source & Text templates \\
        \midrule
         & ``This is accelerometer data for a person that is \{activity\_name\}'' \\
         & ``This person is engaged in \{activity\_name\}'' \\
         & ``This is a person that is \{activity\_name\}'' \\
         & ``This is accelerometer data for \{activity\_name\}'' \\
         & ``\{activity\_name\}'' \\
         & ``This is wearable sensor data for a person engaged in \{activity\_name\}'' \\
         & ``This is wearable sensor data for \{activity\_name\}'' \\
       \multirow{-8}{0.1\textwidth}{Hand-crafted templates}  & ``the person is \{activity\_name\}'' \\

        \midrule
         \multirow{25}{0.1\textwidth}{Templates generated by ChatGPT} & ``This accelerometer data captures movements typical for a person engaged in \{activity\_name\}.'' \\ 
         & ``The person in focus is actively involved in \{activity\_name\} according to wearable sensors.'' \\ 
         & ``These readings represent wearable sensor data from a person engaged in \{activity\_name\}.'' \\ 
         & ``The data collected is from wearable sensors depicting \{activity\_name\} engagement.'' \\ 
         & ``This dataset showcases movements aligned with a person performing \{activity\_name\}.'' \\ 
         & ``The wearer's actions match those expected during \{activity\_name\} as per sensor data.'' \\ 
         & ``These sensor readings portray a person actively participating in \{activity\_name\}.'' \\ 
         & ``The captured data illustrates a person's engagement in \{activity\_name\} through wearable sensors.'' \\ 
         & ``This data captures movements indicative of someone involved in \{activity\_name\}.'' \\ 
          & ``These readings from wearable sensors align with a person doing \{activity\_name\}.'' \\ 
         & ``The sensor data reflects a person's actions consistent with \{activity\_name\} engagement.'' \\ 
         & ``The recorded movements correspond to those expected during \{activity\_name\}.'' \\ 
         & ``This dataset represents a person's involvement in \{activity\_name\} as per sensor readings.'' \\ 
         & ``These sensor readings suggest active participation in \{activity\_name\} by the individual.'' \\ 
         & ``The captured data showcases a person's engagement in \{activity\_name\} through wearables.'' \\ 
         
         & ``This is accelerometer data for a person engaged in \{activity\_name\}.'' \\ 
         & ``The individual in this data is performing \{activity\_name\}.'' \\ 
         & ``This is wearable sensor data capturing \{activity\_name\} movements.'' \\ 
         & ``The recorded actions indicate \{activity\_name\} as per sensor readings.'' \\ 
         & ``This data represents \{activity\_name\} activities observed through wearables.'' \\ 
         & ``This is sensor data for someone doing \{activity\_name\}.'' \\ 
         & ``The captured movements correspond to \{activity\_name\} based on sensors.'' \\ 
         & ``This dataset showcases \{activity\_name\} as captured by wearables.'' \\ 
         & ``The wearer's actions match \{activity\_name\} according to sensor data.'' \\ 
          & ``This person is engaged in \{activity\_name\} as indicated by wearables.'' \\
        \bottomrule
    \end{tabular}
    \captionof{table}{List of text templates used to create sentences of activities.}
    \label{tab:list_of_templates}
\end{minipage}
\clearpage

\subsection{Solution: Adapting Projection Layers on Target Data}
\label{sec:sub:adapting_proj}
\medskip
\begin{minipage}{\textwidth}
    \centering
    \includegraphics[width=\textwidth]{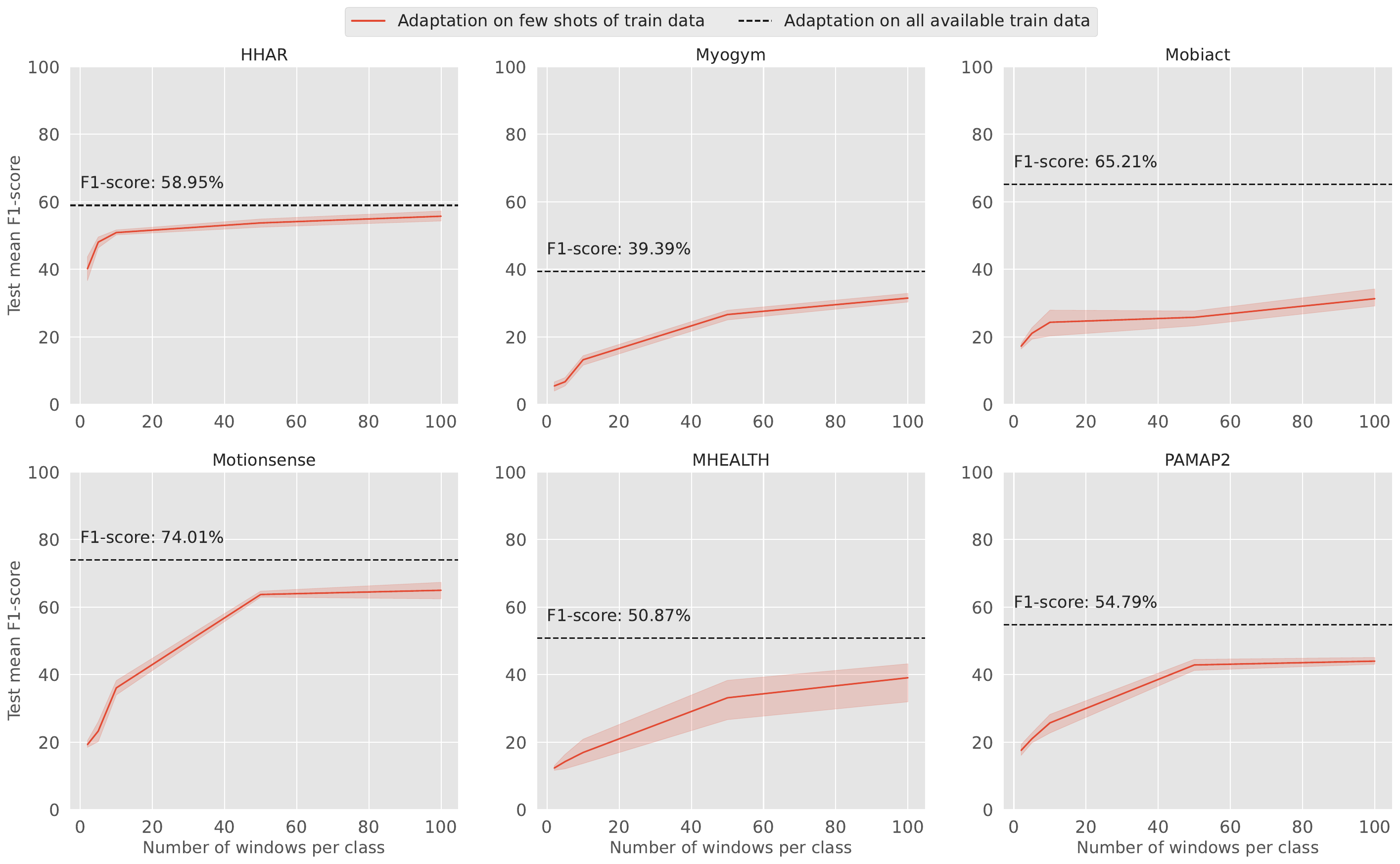}
    \captionof{figure}{  
    Evaluating the impact of adaptation on target data: we observe that access to even small quantities of target data ($<$2 min) substantially improves performance.
    }
    \label{fig:adaptation_few_shot}
\end{minipage}
\clearpage

\subsection{Cross Modal Retrieval of Videos}
\medskip
\begin{minipage}{\textwidth}
    \centering
    \includegraphics[width=\textwidth]{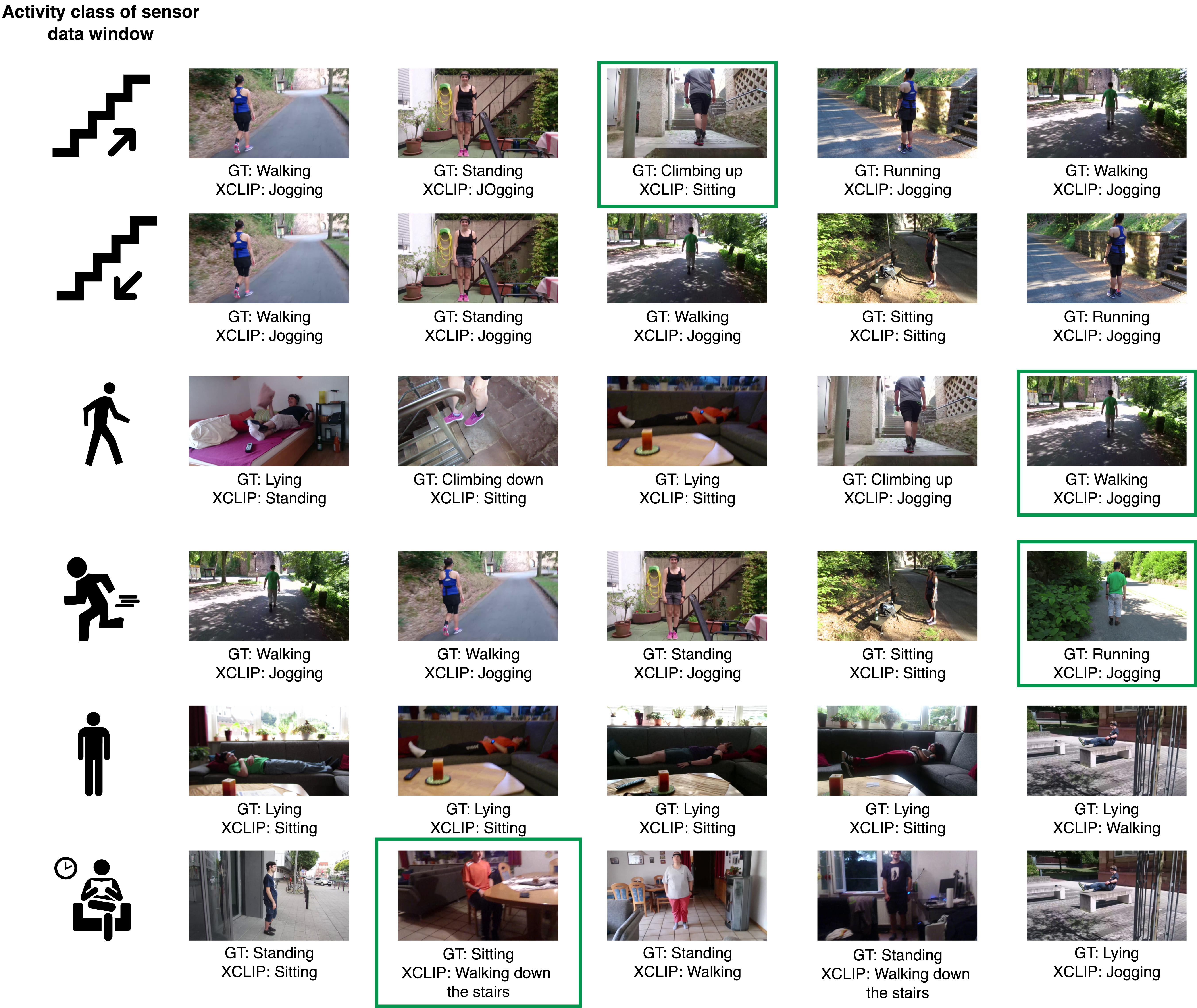}
    \captionof{figure}{Evaluating the cross modal retrieval capabilities of sensor-language pre-training: we observe that for four of the six activities from Motionsense, correct videos are retrieved among the top-5 matches.
    This indicates the potential of sensor-language pre-training, as it can facilitate search functionality across modalities, even ones that are not seen during training (e.g., videos).
    In this figure, `GT' corresponds to the ground truth label from the RealWorld dataset, whereas `XCLIP' comprises the activity predictions from the pre-trained X-CLIP model. 
    }
    \label{fig:cross_modal_retrieval}
\end{minipage}